\documentclass[10pt,twocolumn,letterpaper]{article}

\usepackage[pagenumbers]{cvpr} 

\definecolor{cvprblue}{rgb}{0.21,0.49,0.74}
\usepackage[pagebackref,breaklinks,colorlinks,allcolors=cvprblue]{hyperref}

\makeatletter
\usepackage{amsmath} %
\usepackage{graphicx} %
\usepackage[misc]{ifsym}
\def\paperID{3893} 
\def\confName{CVPR}
\def\confYear{2025}

\newcommand{\authornote}[1]{{%
  \let\thempfn\relax
  \footnotetext[0]{#1}
}}

\title{CATANet: Efficient Content-Aware Token Aggregation for Lightweight
Image Super-Resolution}

\author{
 Xin Liu \quad Jie Liu\textsuperscript{\Letter} \quad Jie Tang \quad Gangshan Wu \\ 
State Key Laboratory for Novel Software Technology, Nanjing University, Nanjing 210023, China \\
{\tt\small xinliu2023@smail.nju.edu.cn \quad \{liujie,tangjie,gswu\}@nju.edu.cn}\\
{\small \url{https://github.com/EquationWalker/CATANet}}
}

\usepackage{algorithm}
\usepackage{algpseudocode}
\usepackage{multirow}
\usepackage{arydshln}
\usepackage{wrapfig, blindtext} 
\usepackage{adjustbox}
\usepackage{amssymb}
\usepackage{utfsym}
\algnewcommand{\algorithmicgoto}{\textbf{go to}}%
\algnewcommand{\Goto}[1]{\algorithmicgoto~\ref{#1}}
\DeclareMathOperator{\F}{F}

\DeclareMathOperator{\MSA}{MSA}
\DeclareMathOperator{\pushback}{PushBack}

\newcommand{\X}{\mathbf{X}}
\newcommand{\C}{\mathbf{C}}

\newcommand{\Q}{\mathbf{Q}}
\newcommand{\K}{\mathbf{K}}
\newcommand{\V}{\mathbf{V}}
\newcommand{\W}{\mathbf{W}}

\begin{document}
\maketitle
\authornote{\Letter: Corresponding author (liujie@nju.edu.cn).}

\begin{abstract}
Transformer-based methods have demonstrated impressive performance in low-level visual tasks such as Image Super-Resolution (SR). However, its computational complexity grows quadratically with the spatial resolution. A series of works attempt to alleviate this problem by dividing Low-Resolution images into local windows, axial stripes, or dilated windows. SR typically leverages the redundancy of images for reconstruction, and this redundancy appears not only in local regions but also in long-range regions. However, these methods limit attention computation to content-agnostic local regions, limiting directly the ability of attention to capture long-range dependency. To address these issues, we propose a lightweight Content-Aware Token Aggregation Network (CATANet). Specifically, we propose an efficient Content-Aware Token Aggregation module for aggregating long-range content-similar tokens, which shares token centers across all image tokens and updates them only during the training phase. Then we utilize intra-group self-attention to enable long-range information interaction. Moreover, we design an inter-group cross-attention to further enhance global information interaction. The experimental results show that, compared with the state-of-the-art cluster-based method SPIN, our method achieves superior performance, with a maximum PSNR improvement of \textbf{\textit{0.33dB}} and nearly \textbf{\textit{double}} the inference speed.
\end{abstract}
\vspace{-0.5cm}    

\section{Introduction}
\begin{figure}
  \centering

   \includegraphics[width=\linewidth]{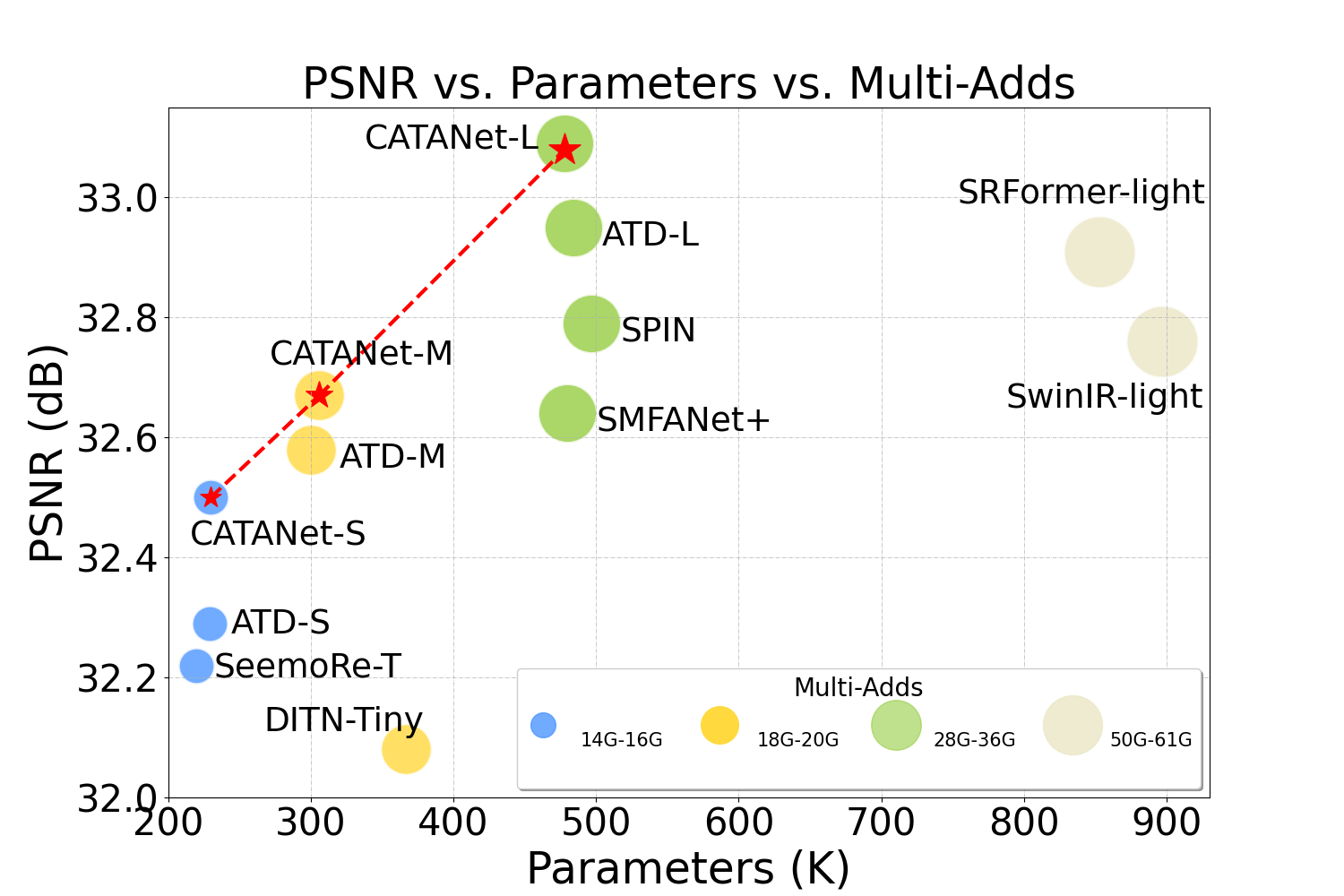}
   \caption{Performance and model complexity comparison on Urban100
dataset for upscaling factor ×2.}
   \label{fig:demo}
   \vspace{-9mm}
\end{figure}
Single Image Super-Resolution (SISR) is a classic task in computer vision and image processing. Its goal is to recover High-Resolution (HR) images from its Low-Resolution (LR) counterpart. SISR is widely applied in various fields, such as medical imaging, digital photography, and reducing server costs for streaming media transmission.\begin{figure*}
  \centering
 \includegraphics[width=\textwidth]{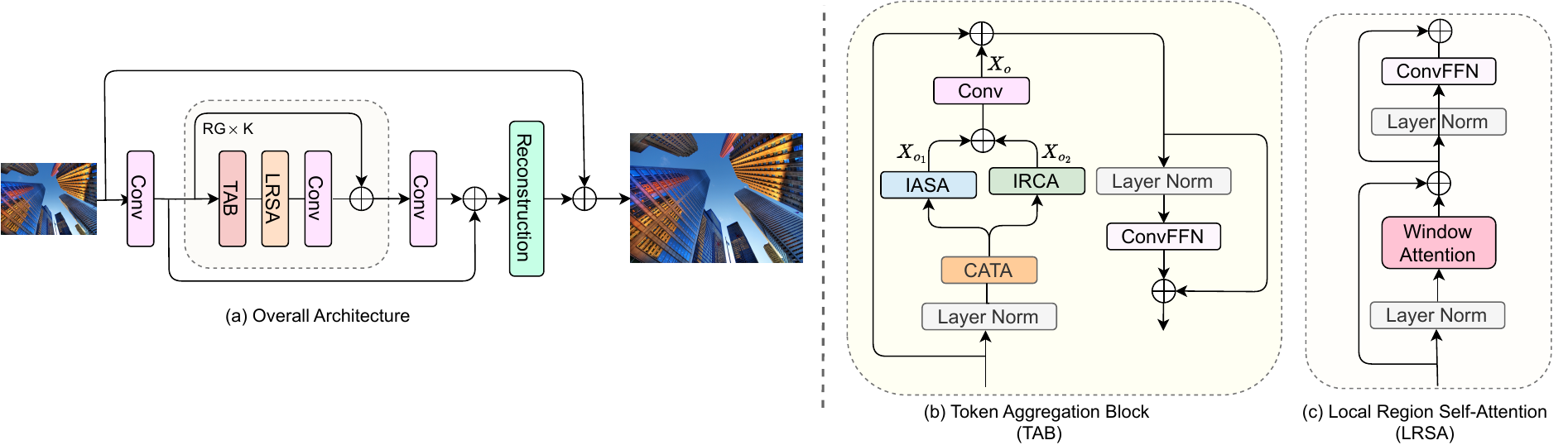}
 \vspace{-6mm}
  \caption{The overall architecture of CATANet and the structure of Token Aggregation Block and Local-Region Self-Attenion.}
  \label{fig:overall}
  \vspace{-0.6cm}
\end{figure*}

Since the pioneering work of Convolutional Neural Networks (CNNs) in SISR~\cite{Dong_Loy_He_Tang_2014}, numerous CNN-based methods~\cite{Dong_Loy_Tang_2016,Kong_Zhao_Qiao_Dong_2021,Li_Liu_Chen_CAi_Gu_Qiao_Dong,Lim_Son_Kim_Nah_Lee_2017,Zhang_Tian_Kong_Zhong_Fu_2018,Li_Yang_Liu_Yang_Jeon_Wu_2019,Zhang_Zeng_Zhang_2021} have been proposed to challenge the reconstruction of HR images from LR images. Due to the local mechanism of convolution, which limits the capture of global dependencies, some CNN-based methods~\cite{Lim_Son_Kim_Nah_Lee_2017,Zhang_Tian_Kong_Zhong_Fu_2018} use very deep and complex network architectures to increase receptive fields and achieve better performance. However, these methods inevitably increase computational resources, which restricts their applicability.

Recently, with the success of transformers in NLP~\cite{vaswani2017attention},  transformers have been applied to multiple high-level computer vision tasks~\cite{dosovitskiy2020image,ding2022davit,liu2021swin,Wang_Xie_Li_Fan_Song_Liang_Lu_Luo_Shao_2021,Chen_Wu_Wang_Hu_Hu_Ding_Cheng_Wang_Vis}, achieving remarkable results. The key success of transformer-based approaches is the Self-Attention mechanism, which can effectively capture long-range dependencies. Due to the powerful potential of transformers, they have also attracted attention in low-level computer vision tasks~\cite{liang2021swinir,Zamir_Arora_Khan_Hayat_Khan_Yang_2022,Wang_Cun_Bao_Zhou_Liu_Li_2022,Chen_Zhang_Gu_Zhang_Kong_Yuan,wang2023omni,Zhang_Zeng_Guo_Zhang,chen2023activating}, including image super-resolution. HAT~\cite{chen2023activating} shows through experiments that utilizing more global information can effectively enhance the quality of reconstructed images. However, these methods divide larger images into smaller local regions for separate processing to alleviate the high complexity of global self-attention.  Although this strategy can improve the efficiency of transformer-based models and provide more refined local information, it still has some limitations. SwinIR~\cite{liang2021swinir} divides the image into content-agnostic local windows, limiting the use of similar tokens over long ranges and resulting in undesirable results. Axial stripe attention~\cite{dong2022cswin} expands the receptive field in a cross-shaped pattern, but this approach remains content-agnostic and may introduce some irrelevant interfering information.

To alleviate these issues, some methods have explored clustering-based solutions.  SPIN~\cite{zhang2023lightweight}, for instance, employs the soft k-means-based token algorithm \cite{Jampani_Sun_Li_Yang_Kautz_2018} for clustering, using the cluster centers as proxies between query and key in the attention mechanism, facilitating the propagation of long-range information. However, SPIN still encounters two primary limitations: (1) The cluster centers provide a sparse representation of the image tokens. However, relying exclusively on these centers for long-range information propagation results in a coarse approximation, which is insufficient for capturing and leveraging detailed long-range dependencies. (2) The inference speed of SPIN is limited due to the need for iterative processing over clustering centers during inference, which constrains the deployment of lightweight models. ATD~\cite{zhang2024transcending} introduces an auxiliary dictionary to learn priors from the training data and uses the dictionary to classify tokens, leading to more accurate token grouping. However, ATD applies multiple attention mechanisms concurrently to boost performance, which significantly increases computational burden and makes ATD less suitable for lightweight scenarios, as shown in Fig.\ref{fig:demo}.

To address these challenges, we propose a novel efficient Content-Aware Token Aggregation Network (CATANet) with Token-Aggregation Block as its core component. Token-Aggregation Block mainly consists of efficient Content-Aware Token Aggregation module, Intra-Group Self-Attention, and Inter-Group Cross-Attention. In contrast to the clustering methods previously discussed, we designed an efficient token centers updating strategy within our Content-Aware Token Aggregation (CATA) module. It shares token centers across all image tokens and updates them only during the training phase, eliminating the impact of updating token centers on model inference speed. Unlike SPIN~\cite{zhang2023lightweight}, which uses clustering centers for long-range information propagation, our Intra-Group Self-Attention employs CATA module to efficiently aggregate content-similar tokens together, forming content-aware regions. Attention is performed within each group, allowing for finer-grained long-range interactions among token information. Moreover, we introduce Inter-Group Cross-Attention, which applies cross-attention between each group and token centers, further enhancing global information interaction.

In summary, our main contributions are as follows:\\
$\bullet$ We propose a novel lightweight image SR network, Content-Aware Token Aggregation Network (CATANet). Our CATANet combines token aggregation with attention mechanisms to capture long-range dependencies while ensuring high inference efficiency.\\
$\bullet$ To mitigate the impact of token centers updates on inference speed, we designed efficient Content-Aware Token Aggregation (CATA) that only updates the token centers during the training phase.\\
$\bullet$ We propose Intra-Group Self-Attention and Inter-Group Cross-Attention, which operate between tokens and effectively capture long-range and global dependencies, effectively reduce interference from irrelevant information, and achieve computational complexity comparable to local-region-based methods.\\
$\bullet$ We conducted extensive experiments to demonstrate that our method surpasses the state-of-the-art cluster-based lightweight SR method SPIN, with a maximum PSNR improvement of 0.33 dB and nearly double the inference speed.

\section{Related Work}
\begin{figure*}
  \centering
 \includegraphics[width=0.8\linewidth]{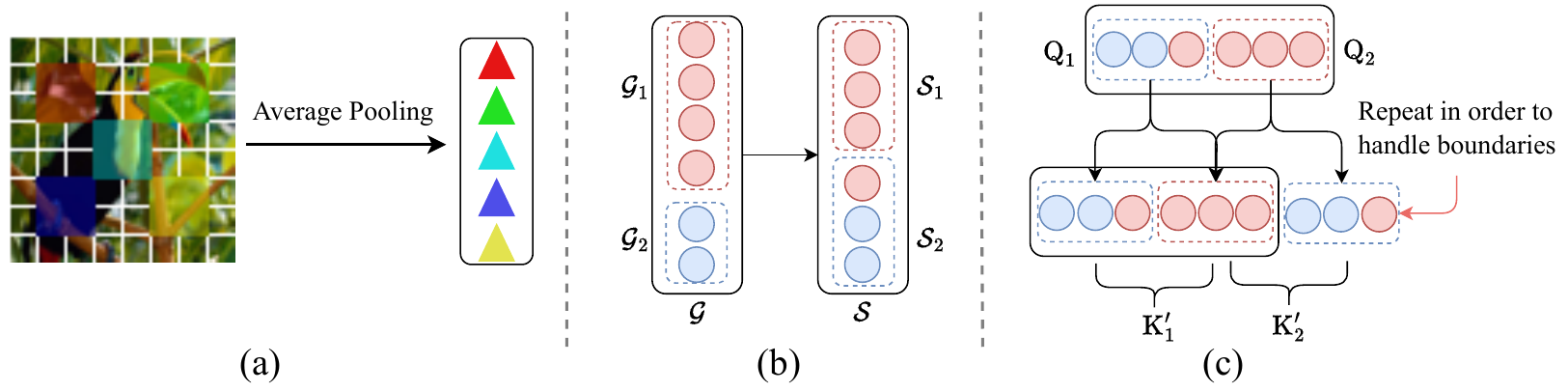}
 \vspace{-0.2cm}
  \caption{(a) A simple illustration for obtaining the initial token centers. (b) Visualization of sub-grouping. The dashed boxes of the same color indicate the same group (left) or subgroup (right). (c) Each subgroup's $\Q_j$ to attend to the $\K_j$ of two consecutive subgroups,  where the same color denotes the same group $\mathcal{G}_j$}
  \label{fig:th}
  \vspace{-6mm}
\end{figure*}
\textbf{Deep Networks for SR.} Deep neural networks have become the mainstream solutions for Image Super-Resolution in recent years due to their powerful representation learning capabilities. Since SRCNN \cite{Dong_Loy_He_Tang_2014} first successfully applied CNNs to the SR field through a three-layer CNN network, a large number of CNN-based methods have achieved state-of-the-art performance using more complex or efficient structures, such as those with residual connections~\cite{Lim_Son_Kim_Nah_Lee_2017,Zhang_Tian_Kong_Zhong_Fu_2018,Wang_Yu_Wu_Gu_Liu_Dong_Qiao_Loy_2019,liu2020residual} and U-shaped architectures~\cite{Cheng_Matsune_Li_Zhu_Zang_Zhan_2019,Mao_Shen_Yang_2016,Liu_Zhang_Zhang_Lin_Zuo_2018}. Compared to CNNs, attention mechanisms have better long-range modeling capabilities, so attention mechanisms have also been introduced into SR to extract the most important features over long ranges. For example, RACN~\cite{Zhang_Li_Li_Wang_Zhong_Fu_2018} utilizes channel attention, while CSFM~\cite{Hu_Li_Huang_Gao_2020} and DAT~\cite{chen2023dual} combines spatial and channel attention. Recently, a series of transformer-based methods have been proposed and have refreshed the state-of-the-art (SOTA), demonstrating the powerful representation learning ability of transformers. For example, SwinIR \cite{liang2021swinir} applies the Swin Transformer \cite{liu2021swin} framework to SR by dividing the entire image into small windows of size $8\times 8$ and shifting the windows when applying multi-head attention mechanisms. Although the above methods can efficiently extract informative features, they often require a large number of parameters. \\
\textbf{Lightweight Super-Resolution Methods.} Due to the urgent demands for applying networks to resource-constrained devices, lightweight SR has attracted widespread attention\cite{li2022blueprint,Dong_Loy_He_Tang_2014,Shi_Caballero_Huszar_Totz_Aitken_Bishop_Rueckert_Wang_2016,zamfir2024details,smfanet}. For example, ESPCN \cite{Dong_Loy_He_Tang_2014} and FSRCNN \cite{Shi_Caballero_Huszar_Totz_Aitken_Bishop_Rueckert_Wang_2016} utilize post-upsampling techniques to reduce computational costs. IDN \cite{hui2018fast} and IMDN \cite{Hui_Gao_Yang_Wang_2019} employ information distillation blocks to extract useful information, dividing the input features into two streams using slicing operations, with one stream further enhancing feature representation through convolution. Subsequently, the two features are combined to obtain richer information. For lightweight super-resolution based on transformers, most methods focus on sparse attention mechanisms, such as window-based attention \cite{liang2021swinir} and group-wise attention \cite{Zhang_Zeng_Guo_Zhang}, to reduce the high computational complexity of attention calculation. Although these methods effectively reduce computational complexity, their sparse nature is content-agnostic and cannot guarantee the quality of SR reconstruction.
\\ \textbf{Token Clustering for Computer Vision.} Token clustering is a well-studied task in computer vision, and recent advances in deep learning have made significant progress in this area. A common approach is to use traditional clustering algorithms, such as k-means, to gather similar tokens together. For example, BOAT \cite{yu2022boat} employs hierarchical clustering to gather similar tokens together and computes attention within each cluster. SPIN \cite{zhang2023lightweight} utilizes the soft k-means-based token algorithm \cite{Jampani_Sun_Li_Yang_Kautz_2018} to aggregate tokens, using the aggregated tokens as proxies between Query and Key to leverage long-range information. Although these token clustering methods have shown promising results in computer vision, the computational cost of the clustering process is significant, which hinders the application of token clustering in lightweight SR.
\section{Method}

\subsection{Network Architecture}
The overall network of the proposed CATANet comprises three modules: shallow feature extraction, deep feature extraction, and image reconstruction, as illustrated in Fig.~\ref{fig:overall}(a). The shallow feature extraction is performed by a $3\times 3$ convolutional layer, which maps the image from the original input space to a high-dimensional feature space. Let $\mathbf{I}_{\text{LR}}$ and $\mathbf{I}_{\text{SR}}$ represent the Low-Resolution (LR) input image and the Super-Resolved (SR) image, respectively. We first get shallow feature $\X_0$ by
\begin{equation*}
\X_0=\F_{\text{S}}(\mathbf{I}_{\text{LR}}),
\end{equation*}
where $\F_\text{S}$ denotes the function of the shallow feature extraction. Then deeper features are extracted by sequential Residual Group (RG) of $\text{K}$. Each RG includes three components: Token-Aggregation Block (TAB), Local-Region Self-Attention (LRSA), and $3\times3$ convolution (Conv).

The input features of each RG are first processed through a TAB module to perform token aggregation, capturing long-range dependencies. Then, we use a LRSA to enhance the dependencies between tokens in local regions. Furthermore, we employ a $3\times3$ convolution at the end of each RG to further refine local features and implicitly learn positional embedding. Formally, for the $i$-th RG, the whole process can be formulated as:
\begin{equation*}
\X_i=\X_{i-1}+\F_{\text{Conv}}\left(\F_{\text{LRSA}}\left(\F_{\text{TAB}}\left(\X_{i-1}\right)\right)\right),
\end{equation*}
where $\X_i$ denotes the output feature in the $i$-th RG and $\F(\cdot)$ denotes the function of each individual component. Following previous work~\cite{liang2021swinir,liu2020residual}, residual connections are used to stabilize the training process.

After $\text{K}$ RGs, we use the image reconstruction module to obtain global residual information, which is then added to the upsampled image of $\mathbf{I}_{\text{LR}}$ to obtain the high-resolution image $\mathbf{I}_{\text{SR}}$.
\begin{equation*}
\mathbf{I}_{\text{SR}}=\F_{\text{Up}}(\mathbf{I}_{\text{LR}})+\F_{\text{IR}}(\X_K),
\end{equation*}
where $\F_{\text{Up}}$ denotes the function of the upsampling operator and $\F_{\text{IR}}$ denotes the function of the image reconstruction  that includes a $3\times 3$ convolution and a pixel shuffle~\cite{Shi_Caballero_Huszar_Totz_Aitken_Bishop_Rueckert_Wang_2016}.

\subsection{Token-Aggregation Block (TAB)}
In previous token clustering-based methods, BOAT~\cite{yu2022boat} proposed a hierarchical clustering-based attention mechanism to achieve fine-grained long-range information interaction. However, the hierarchical clustering requires iterative updates of cluster centroids, slowing down model inference. SPIN~\cite{zhang2023lightweight} uses cluster centers as proxies between Query and Key in self-attention, facilitating long-range information propagation. This method, though, results in coarse information transmission, inevitably introducing irrelevant information. In contrast, we propose Intra-Group Self-Attention to achieve more refined information interaction between content-similar tokens. Moreover, like BOAT, SPIN suffers from slow inference speed.

To address these issues more effectively, we propose Token-Aggregation Block (TAB). As shown in Fig.~\ref{fig:overall}(b), TAB mainly consists of four parts: Content-Aware Token Aggregation module, Intra-Group Self-Attention, Inter-Group Cross-Attention and a 1$\times$1 convolution. By utilizing these four modules, we can efficiently  achieve fine-grained long-range information interaction while allowing us to enjoy computational complexity similar to local-region attention. 
\subsubsection{Content-Aware Token Aggregation (CATA)}
\algnewcommand{\NoNumber}[1]{\Statex \hspace{0em} \(\triangleright\) #1}
\begin{algorithm}
\caption{Content-Aware Token Aggregation}
\label{Alg:ECTA}
\begin{algorithmic}[1]
\NoNumber \textbf{input :}  image tokens $\X=\{x_i\in\mathbb{R}^{d}\}_{i=1\cdots N}$
\NoNumber \textbf{input :} token centers $\C=\{c_j\in\mathbb{R}^{d}\}_{j=1\cdots M}$
\NoNumber \textbf{output :} token groups $\{\mathcal{G}_j\}_{j=1\cdots M}$ 
\If{\textbf{not training}}
\State \Goto{marker}
\EndIf
\State $\{c_j^\prime\}=\{c_j\}$
\For{$t = T, \dots, 1$}
\State \textcolor{gray}{// the similarity between tokens and token centers.}
\State $\mathbf{D} = \mathcal{M}(\{x_i\}, \{c_j^\prime\})$ \textcolor{gray}{// $\mathbf{D}\in\mathbb{R}^{N\times M}$}
\State $\{\mathcal{G}_j\}=\left\{\left\{x^i\big|\arg\max_{k}\mathbf{D}_{ik}=j\right\}\right\}$
\State $\{c_j^\prime\}=\left\{\frac{1}{|\mathcal{G}_j|}\sum_{k=1}^{|\mathcal{G}_j|}x_k\big|x_k\in\mathcal{G}_j\right\}$
\EndFor
\State \textcolor{gray}{// Update token centers using EMA.}
\State $\{c_j\}= \{\lambda\cdot c_j+(1-\lambda)\cdot c_j^\prime\}$
\State $\mathbf{D} = \mathcal{M}(\{x_i\}, \{c_j\})$ \label{marker}
\State $\{\mathcal{G}_j\}=\left\{\{x^i\big|\arg\max_{k}\mathbf{D}_{ik}=j\}\right\}$
\end{algorithmic}
\end{algorithm}
The process of CATA module is illustrated in \cref{Alg:ECTA}, where $\mathcal{M}$ is the cosine similarity function. As shown in \cref{fig:th}(a), we obtain the initial $M$ token centers $\{c_j\}_{j=1\cdots M}$ by simply performing average pooling over regular regions. Inspired by Routing Transformer~\cite{roy2021efficient}, we share token centers among all image tokens, aiming to learn a set of global token centers across the training dataset rather than for each individual image. We update token centers using Exponential Moving Average (EMA), with a decay parameter $\lambda$ typically set to 0.999.

\begin{figure*}[t]
    \centering
    
    \captionsetup{font={small}}
    \begin{subfigure}{.19\linewidth}
        \centering
        \includegraphics[width=\linewidth]{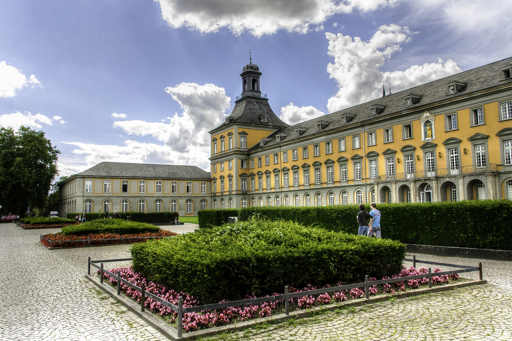}
        \caption{}
    \end{subfigure}%
    \hfill
    \begin{subfigure}{.19\linewidth}
        \centering
        \includegraphics[width=\linewidth]{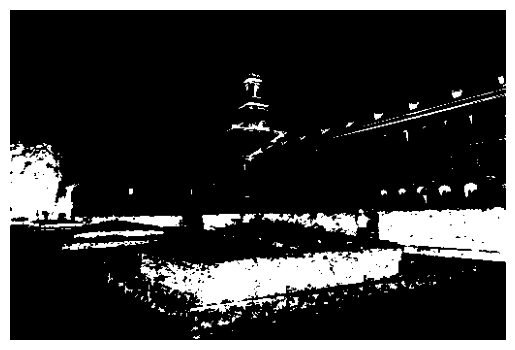}
        \caption{}
    \end{subfigure}
    \hfill
    \begin{subfigure}{.19\linewidth}
        \centering
        \includegraphics[width=\linewidth]{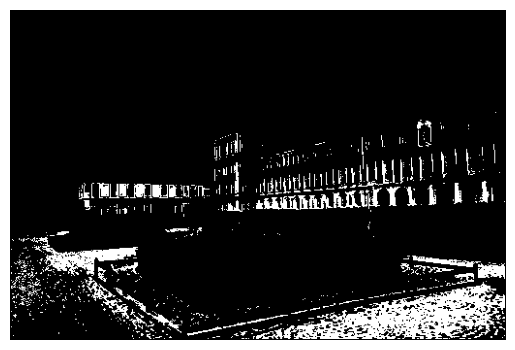}
        \caption{}
    \end{subfigure}%
    \hfill
    \begin{subfigure}{.19\linewidth}
        \centering
        \includegraphics[width=\linewidth]{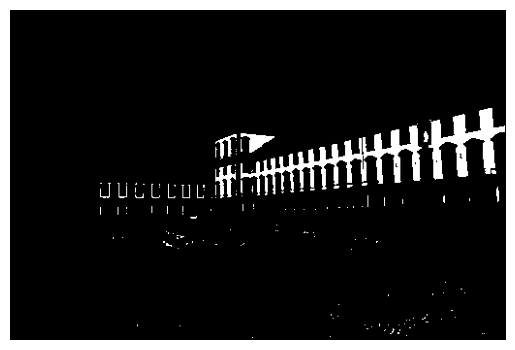}
        \caption{}
    \end{subfigure}
    \hfill
    \begin{subfigure}{.19\linewidth}
        \centering
        \includegraphics[width=\linewidth]{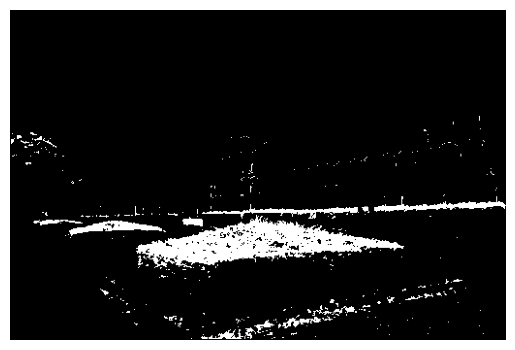}
        \caption{}
    \end{subfigure}
\caption{Visualization of token grouping results of TAB. (a) is the input image. The white part of each binarized image from (b) - (e) represents a single token group $\mathcal{G}_j$.}
\label{fig:vis_tab}
\vspace{-0.5cm}
\end{figure*}

Image tokens are divided into \( M \) content-similar token groups based on the similarity between each token and the token centers. A visualized example of the CATA module is shown in Fig.~\ref{fig:vis_tab}, where it can be seen that our token groups aggregate content-similar tokens over long ranges, resulting in more precise token grouping. However, the number of tokens in each
group may differ, which results in low parallelism efficiency. To address the issue of unbalanced group, as shown in Fig.~\ref{fig:th}(b), we refer to~\cite{mei2021image} to further divide the groups $\mathcal{G}$ into subgroups $\mathcal{S}$:
\begin{table}
\vspace{-0.3cm}
\footnotesize
\caption{Ablation study attending to consecutive subgroups. PSNR are calculated with a scale factor of 4. }
\vspace{-6mm}
\begin{center}
\resizebox{\linewidth}{!}{
\begin{tabular}{|l|c|c|c|c|}
\hline
Method & Params& Multi-Adds & Urban100& Manga109\\
\hline
Not Attend& 536K&46.8G& 26.85&31.26\\
\textbf{Attend (ours)}& 536K& 46.8G & \textcolor{red}{26.87} & \textcolor{red}{31.31}\\
\hline
\end{tabular}}
\label{table:qk}
\end{center}
\vspace{-10mm}
\end{table}
\begin{equation*}
    \begin{aligned}
\mathcal{G}&=\left[\mathcal{G}_1^{1},\mathcal{G}_1^{2}\cdots,\mathcal{G}_1^{n_1},\cdots,\mathcal{G}_{M}^{n_M}\right],\\
\mathcal{S}_j&=\left[ \mathcal{G}_{j * g_s + 1}, \mathcal{G}_{j * g_s + 2}, \cdots, \mathcal{G}_{(j+1) * g_s} \right],\\ 
\mathcal{S}&=\left[\mathcal{S}_1,\mathcal{S}_2,\cdots,\mathcal{S}_j,\cdots\right],
    \end{aligned}
\end{equation*}
where the group $\mathcal{G}_i$ contains $n_i$ tokens. Each group is concatenated to form $\mathcal{G}$, then divided into subgroups $\mathcal{S}$. After division, all subgroups have the same fixed size $g_s$, improving parallelism efficiency. As shown in Tab.~\ref{tab:time}(last two rows), by utilizing sub-grouping, the inference speed of CATANet is approximately doubled.
\subsubsection{Intra-Group Self-Attention (IASA)}
Given the subgroups $\mathcal{S}=\left\{\mathcal{S}_j\right\}$, we project it into matrix $\{\Q_j\}$, $\{\K_j\}$, and $\{\V_j\}$, as follows:
\begin{equation*}
\Q_j,\K_j,\V_j=\mathcal{S}_j\W^Q,\mathcal{S}_j\W^K,\mathcal{S}_j\W^V,
\end{equation*}
where $\W^Q$,$\W^K$ and $\W^V\in\mathbb{R}^{d\times d}$ are weight matrices. To enhance parallel efficiency, we divide the groups $\mathcal{G}$ into sub-groups $\mathcal{S}$. However, this may result in content-similar tokens being split into adjacent subgroups. To mitigate this issue, we allow each subgroup's $\Q_j$ to attend to the $\K_j$ of two consecutive subgroups, as shown in Fig.\ref{fig:th}(c). Furthermore, as shown in Tab.~\ref{table:qk}, attending to adjacent subgroups effectively enhances performance without adding computational overhead. The procedure can be formulated as:
\begin{equation}
\label{eq:intra}
\begin{aligned}
\K_j^\prime&=\left[\K_{j},\K_{j+1}\right],
\V_j^\prime=\left[\V_{j},\V_{j+1}\right],\\
\mathbf{O}_j&=\MSA\left(\Q_j,\K_j^\prime,\V_j^\prime\right),\\
\X_{o_1} &= \pushback\left(\left\{\mathbf{O}_j\right\}\right),
\end{aligned}
\end{equation}
where  $\MSA(\cdot)$ denotes Multi-Head Self-Attention operation~\cite{vaswani2017attention} and $\pushback$ operation  put each token back to its original position on the feature map to form $\X_{o_1}$.
\subsubsection{Inter-Group Cross-Attention (IRCA)}
As shown in Algorithm~\ref{Alg:ECTA}, during the training phase, the token centers $\C$ aggregate content-similar tokens, summarizing global prior information. To leverage this global prior, we perform cross-attention between each subgroup $\mathcal{S}_j\in\mathbb{R}^{g_s\times d}$ and $\C\in\mathbb{R}^{M\times d}$, as follows:
\begin{equation*}
\begin{aligned}
\mathbf{O}_j^\prime &= \MSA\left(\mathcal{S}_j\W^q,\C\W^k,\C\W^v\right),\\
\X_{o_2} &= \pushback\left(\left\{\mathbf{O}_j^\prime\right\}\right),
\end{aligned}
\end{equation*}
where $\W^q$,$\W^k$ and $\W^v\in\mathbb{R}^{d\times d}$ are weight matrices. We set \( M \ll N \) to maintain a low computational cost. Finally, $\X_o$ is obtained by fusing $\X_{o_2}$ and $\X_{o_1}$ from \cref{eq:intra} through a convolution:
\begin{equation*}
\X_o=\F_{\text{Conv}}\left(\X_{o_1}+\X_{o_2}\right).
\end{equation*}
\subsection{Local-Region Self-Attention (LRSA)}
As shown in Fig.~\ref{fig:overall}(c), the LASA follows HPINet~\cite{liu2023coarse}, using overlapping patches to enhance feature interaction LRSA module is responsible for learning finer local details. Given input $\X_o\in\mathbb{R}^{N\times d}$, this process can be expressed as:
\begin{equation*}
\X_{out}=\MSA(\X_o\W_Q,\X_o\W_K,\X_o\W_V),
\end{equation*}
where $\X_{out}\in\mathbb{R}^{N\times d}$. $\W_Q, \W_K$ and $\W_V$ are weight matrices that are shared across patches.
\vspace{-0.3cm}
\paragraph{ConvFNN.} Similar to the Transformer layer, after TAB and LRSA, ConvFNN~\cite{zhou2023srformer} is employed to perform feature interactions along the channel dimension, as illustrated in Fig.~\ref{fig:overall}. Additionally, layer normalization is added before the TAB, LRSA, and ConvFFN, and residual shortcuts after both modules are added as well.

\begin{table*}[t]
\scriptsize
\caption{Comparison (PSNR/SSIM) with SOTA methods for image SR. Best and second best results are colored with \textcolor{red}{red} and \textcolor{blue}{blue}. $\dagger$ indicate that self-ensemble strategy~\cite{lim2017enhanced} is used in testing.}
\vspace{-0.6cm}
\begin{center}
\resizebox{\linewidth}{!}{
\begin{tabular}{|l|c|c|c|c|c|c|c|c|c|c|c|c|}
\hline
\multirow{2}{*}{Method} & \multirow{2}{*}{Scale} & \multirow{2}{*}{Params} &  \multicolumn{2}{c|}{Set5} & \multicolumn{2}{c|}{Set14} & \multicolumn{2}{c|}{B100} & \multicolumn{2}{c|}{Urban100} & \multicolumn{2}{c|}{Manga109}\\
\cline{4-13}
&& & PSNR & SSIM & PSNR & SSIM & PSNR & SSIM & PSNR & SSIM & PSNR & SSIM
\\
\hline
CARN~\cite{Ahn_Kang_Sohn_2018} & $\times$2 & 1592K&37.76 & 0.9590 & 33.52 & 0.9166 & 32.09 & 0.8978 & 31.92 & 0.9256 & 38.36 & 0.9765\\
IMDN \cite{Hui_Gao_Yang_Wang_2019}& $\times$2 &
694K & 38.00&0.9605 & 33.63&0.9177 & 32.19&0.8996 & 32.17&0.9283 & 38.88&0.9774\\
RFDN~\cite{liu2020residual}&$\times$2&530K&38.05&0.9606&33.68&0.9184&32.16&0.8994&32.12&0.9278&38.88&0.9773\\
AWSRAN-M \cite{wang2019lightweight}& $\times$2 &
1063K&38.04&0.9605&33.66&0.9181&32.21&0.9000&32.23&0.9294&38.66&0.9772\\
OSFFNet \cite{wang2024osffnet}& $\times$2 &
516K&38.11&0.9610&33.72&0.9190&32.29&0.9012&32.67&0.9331&39.09&0.9780\\
ESRT \cite{Lu_Li_Liu_Huang_Zhang_Zeng_2021}& $\times$2 &677K&38.03&0.9600&33.75&0.9184&32.25&0.9001&32.58&0.9318&39.12&0.9774\\
ELAN-light \cite{Zhang_Zeng_Guo_Zhang}& $\times$2& 582K&38.17&0.9611&33.94&0.9207&32.30&0.9012&32.76&0.9340&39.11&0.9782\\
A-CubeNet \cite{hang2020attention}&$\times$2&1380K&38.12&0.9609&33.73&0.9191&32.26&0.9007&32.39&0.9308&38.88&0.9776\\
SwinIR-light~\cite{liang2021swinir} & $\times$2 &  
878K&38.14&0.9611&33.86&0.9206&32.31&0.9012&32.76&0.9340&39.12&0.9783\\
SPIN~\cite{zhang2023lightweight} &$\times$2&497K&38.20&0.9615&33.90&0.9215&32.31&0.9015&32.79&0.9340&39.18&0.9784\\
\hdashline
CATANet (ours) & $\times$2 & 477K& 
\textcolor{blue}{38.28} & \textcolor{blue}{0.9617} & \textcolor{blue}{33.99} & \textcolor{blue}{0.9217} & \textcolor{blue}{32.37} & \textcolor{blue}{0.9023} & \textcolor{blue}{33.09} & \textcolor{blue}{0.9372} & \textcolor{blue}{39.37} & \textcolor{blue}{0.9784}
\\
CATANet$\dagger$ (ours) & $\times$2 & 477K& 
\textcolor{red}{38.35} & \textcolor{red}{0.9620} & \textcolor{red}{34.11} & \textcolor{red}{0.9229} & \textcolor{red}{32.41} & \textcolor{red}{0.9027} & \textcolor{red}{33.33} & \textcolor{red}{0.9387} & \textcolor{red}{39.57} & \textcolor{red}{0.9788}
\\
\hline
CARN~\cite{Ahn_Kang_Sohn_2018} & $\times$3 & 1592K&34.29&0.9255&30.29&0.8407&29.06&0.8034&28.06&0.8493&33.43&0.9427\\
IMDN~\cite{Hui_Gao_Yang_Wang_2019}& $\times$3 &
703K&34.36&0.9270&30.32&0.8417&29.09&0.8046&28.17&0.8519&33.61&0.9445\\
RFDN~\cite{liu2020residual}&$\times$3&540K&34.41&0.9273&30.34&0.8420&29.09&0.8050&28.21&0.8525&33.67&0.9449\\
AWSRAN-M \cite{wang2019lightweight}& $\times$3 &
1143K&34.42&0.9275&30.32&0.8419&29.13&0.8059&28.26&0.8545&33.64&0.9450\\
OSFFNet \cite{wang2024osffnet}& $\times$3 &
524K&34.58&0.9287&30.48&0.8450&29.21&0.8080&28.49&0.8595&34.00&0.9472\\
ESRT \cite{Lu_Li_Liu_Huang_Zhang_Zeng_2021}& $\times$3 &770K&34.42&0.9268&30.43&0.8433&29.15&0.8063&28.46&0.8574&33.95&0.9455\\
ELAN-light \cite{Zhang_Zeng_Guo_Zhang}& $\times$3& 590K&34.64&0.9288&30.55&0.8463&29.21&0.8081&28.69&0.8624&34.00&0.9478\\
A-CubeNet \cite{hang2020attention}&$\times$3&1560K&34.53&0.9281&30.45&0.8441&29.17&0.8068&28.38&0.8568&33.90&0.9466\\
SwinIR-light~\cite{liang2021swinir} & $\times$3 &  
886K&34.62&0.9289&30.54&0.8463&29.20&0.8082&28.66&0.8624&33.98&0.9478\\
SPIN~\cite{zhang2023lightweight} &$\times$3&569K&34.65&0.9293&30.57&0.8464&29.23&0.8089&28.71&0.8627&34.24&0.9489\\
\hdashline
CATANet (ours) & $\times$3 & 550K& 
\textcolor{blue}{34.75} & \textcolor{blue}{0.9300} & \textcolor{blue}{30.67} & \textcolor{blue}{0.8481} & \textcolor{blue}{29.28} & \textcolor{blue}{0.8101} & \textcolor{blue}{29.04} & \textcolor{blue}{0.8689} & \textcolor{blue}{34.40} & \textcolor{blue}{0.9499}
\\
CATANet$\dagger$ (ours) & $\times$3 & 550K& 
\textcolor{red}{34.83} & \textcolor{red}{0.9307} & \textcolor{red}{30.73} & \textcolor{red}{0.8490} & \textcolor{red}{29.34} & \textcolor{red}{0.8111} & \textcolor{red}{29.24} & \textcolor{red}{0.8718} & \textcolor{red}{34.69} & \textcolor{red}{0.9512}
\\
\hline
CARN~\cite{Ahn_Kang_Sohn_2018} & $\times$4 & 1592K&32.13&0.8937&28.60&0.7806&27.58&0.7349&26.07&0.7837&30.42&0.9070\\
IMDN \cite{Hui_Gao_Yang_Wang_2019}& $\times$4 &
715K&32.21&0.8948&28.58&0.7811&27.56&0.7353&26.04&0.7838&30.45&0.9075\\
RFDN~\cite{liu2020residual}&$\times$4&550K&32.24&0.8952&28.61&0.7819&27.57&0.7360&26.11&0.7858&30.58&0.9089\\
AWSRAN-M \cite{wang2019lightweight}& $\times$4 &
1520K&32.32&0.8969&28.72&0.7847&27.65&0.7382&26.27&0.7913&30.81&0.9114\\
OSFFNet \cite{wang2024osffnet}& $\times$4 &
537K&32.39&0.8976&28.75&0.7852&27.66&0.7393&26.36&0.7950&30.84&0.9125\\
ESRT \cite{Lu_Li_Liu_Huang_Zhang_Zeng_2021}& $\times$4 &751K&32.19&0.8947&28.69&0.7833&27.69&0.7379&26.39&0.7962&30.75&0.9100\\
ELAN-light \cite{Zhang_Zeng_Guo_Zhang}& $\times$4& 601K&32.43&0.8975&28.78&0.7858&27.69&0.7406&26.54&0.7982&30.92&0.9150\\
A-CubeNet \cite{hang2020attention}&$\times$4&1520K&32.32&0.8969&28.72&0.7847&27.65&0.7382&26.27&0.7913&30.81&0.9114\\
SwinIR-light~\cite{liang2021swinir} & $\times$4 &  
897K&32.44&0.8976&28.77&0.7858&27.69&0.7406&26.47&0.7980&30.92&0.9151\\

SPIN~\cite{zhang2023lightweight} &$\times$4&555K&32.48&0.8983&28.80&0.7862&27.70&0.7415&26.55&0.7998&30.98&0.9156\\
\hdashline
CATANet (ours) & $\times$4 & 535K& 
\textcolor{blue}{32.58} & \textcolor{blue}{0.8998} & \textcolor{blue}{28.90} & \textcolor{blue}{0.7880} & \textcolor{blue}{27.75} & \textcolor{blue}{0.7427} & \textcolor{blue}{26.87} & \textcolor{blue}{0.8081} & \textcolor{blue}{31.31} & \textcolor{blue}{0.9183}
\\
CATANet$\dagger$ (ours) & $\times$4 & 535K& 
\textcolor{red}{32.68} & \textcolor{red}{0.9009} & \textcolor{red}{28.98} & \textcolor{red}{0.7894} & \textcolor{red}{27.80} & \textcolor{red}{0.7437} & \textcolor{red}{27.04} & \textcolor{red}{0.8113} & \textcolor{red}{31.58} & \textcolor{red}{0.9206}
\\
\hline
\end{tabular}}
\label{tab:psnr_ssim_5sets}
\vspace{-0.8cm}
\end{center}
\end{table*}

\section{Experiments}
\subsection{Experimental Settings}
We train the model using DIV2K~\cite{timofte2017ntire}, a high-quality dataset widely used for SR tasks. It includes 800 training images together with 100 validation images. Additionally, we evaluate our model on five commonly used public super-resolution datasets, including Set5 \cite{Bevilacqua_Roumy_Guillemot_Morel_2012}, Set14 \cite{Zeyde_Elad_Protter_2012}, B100~\cite{Martin_Fowlkes_Tal_Malik_2002}, Urban100~\cite{Huang_Singh_Ahuja_2015}, and Manga109~\cite{Matsui_Ito_Aramaki_Fujimoto_Ogawa_Yamasaki_Aizawa_2017}.\\
We use the metrics PSNR and SSIM \cite{Wang_Bovik_Sheikh_Simoncelli_2004} to evaluate our model performance. Following previous work \cite{liang2021swinir,zhou2023srformer}, both metrics are first transformed into the YCbCr color space and then computed on the Y channel. Details of the training procedure and network hyperparameters can be found in the supplementary material.
\subsection{Comparisons with the state-of-the-arts}
\begin{figure*}
\centering
\resizebox{\linewidth}{!}{
\begin{tabular}{cc}
\begin{adjustbox}{valign=t}
\begin{tabular}{c}
\includegraphics[width=0.42\textwidth]{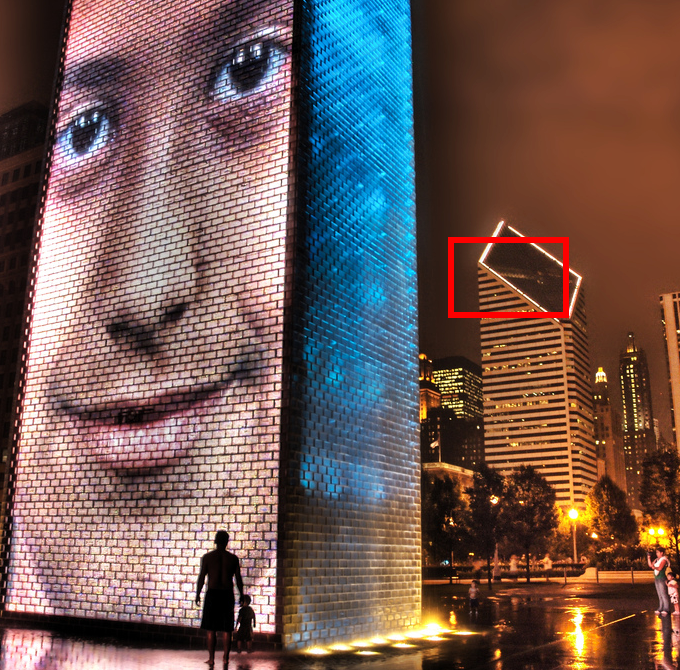}
\\
Urban100: img\_076 ($\times$4)
\end{tabular}
\end{adjustbox}
\begin{adjustbox}{valign=t}
\begin{tabular}{ccccc}
\includegraphics[width=0.285\textwidth]{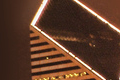} &
\includegraphics[width=0.285\textwidth]{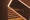}  &
\includegraphics[width=0.285\textwidth]{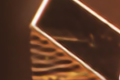} &
\includegraphics[width=0.285\textwidth]{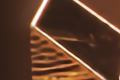}
\\
HQ &
Bicubic  &
A-CubeNet~(25.35/0.8503) &
CARN~(\textcolor{blue}{26.20}/\textcolor{blue}{0.8775})
\\
\includegraphics[width=0.285\textwidth]{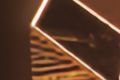}  &
\includegraphics[width=0.285\textwidth]{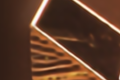}&
\includegraphics[width=0.285\textwidth]{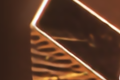}  &
\includegraphics[width=0.285\textwidth]{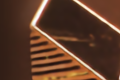}
\\ 
IMDN~(25.66/0.8625) &
ESRT~(25.35/0.8507) &
SwinIR-light~(24.39/0.8327) &
ours~(\textcolor{red}{29.04}/\textcolor{red}{0.9272})
\end{tabular}
\end{adjustbox}\\
\begin{adjustbox}{valign=t}
    \begin{tabular}{c}
    \includegraphics[width=0.42\textwidth]{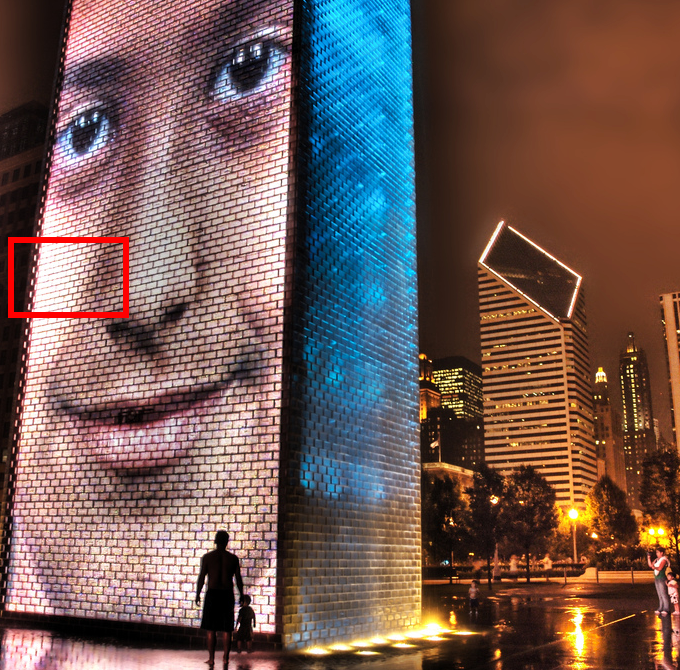}
    \\
    Urban100: img\_076 ($\times$4)
    \end{tabular}
    \end{adjustbox}
    \begin{adjustbox}{valign=t}
    \begin{tabular}{ccccc}
    \includegraphics[width=0.285\textwidth]{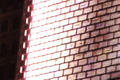} &
    \includegraphics[width=0.285\textwidth]{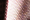}  &
    \includegraphics[width=0.285\textwidth]{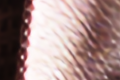} &
    \includegraphics[width=0.285\textwidth]{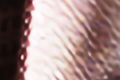}
    \\
    HQ &
    Bicubic  &
    A-CubeNet~(18.46/0.3580) &
    CARN~(18.54/0.3619)
    \\
    \includegraphics[width=0.285\textwidth]{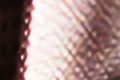}  &
    \includegraphics[width=0.285\textwidth]{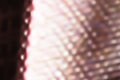}&
    \includegraphics[width=0.285\textwidth]{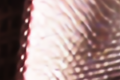}  &
    \includegraphics[width=0.285\textwidth]{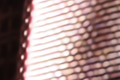}
    \\ 
    IMDN~(19.09/0.4128) &
    ESRT~(\textcolor{blue}{19.88}/\textcolor{blue}{0.4891}) &
    SwinIR-light~(19.37/0.4500) &
    ours~(\textcolor{red}{22.71}/\textcolor{red}{0.7967})
    \\
    \end{tabular}
    \end{adjustbox}
\end{tabular}
}
\vspace{-2mm}
\caption{Visual comparisons of CATANet and other state-of-the-art lightweight SR methods. Metrics (PSNR/SSIM) are calculated on
each patch. Best and second best results are colored with \textcolor{red}{red} and \textcolor{blue}{blue}, respectively.}
\label{fig:vis_compare}
\vspace{-5mm}
\end{figure*}

We compare the proposed CATANet with commonly used lightweight SR models for scaling factors $\times 2, \times 3,$ and $\times 4$, including CNN-based models (CRAN~\cite{Ahn_Kang_Sohn_2018}, LatticeNet~\cite{Luo_Xie_Zhang_Qu_Li_Fu_2020}, IMDN \cite{Hui_Gao_Yang_Wang_2019}, RFDN~\cite{liu2020residual}, AWSRAN-M \cite{wang2019lightweight}, and OSFFNet \cite{wang2024osffnet}) and transformer-based models (ESRT \cite{Lu_Li_Liu_Huang_Zhang_Zeng_2021}, ELAN-light \cite{Zhang_Zeng_Guo_Zhang}, A-CubeNet \cite{hang2020attention}, SwinIR \cite{liang2021swinir}, and SPIN \cite{zhang2023lightweight}). Tab.~\ref{tab:psnr_ssim_5sets} shows the quantitative comparison results of our model and other models in terms of PSNR and SSIM. As shown, our CATANet outperforms the compared methods on all benchmark datasets with three factors. Meanwhile, we have 15K fewer parameters compared to SPIN. It is notable that, thanks to the more refined long-range information interaction, our CATANet$\dagger$ outperforms SPIN by a maximum PSNR of 0.60dB, which is a significant improvement in image SR. Even without self-ensemble, our CATANet still surpass SPIN with a PSNR improvement of 0.33 dB. Additionally, we conduct a more comprehensive and fair comparison with more state-of-the-art methods in Sec.~\ref{sec:mz_time}.\\
\textbf{Visual Comparison.} We provide some visual examples using different methods in Fig.~\ref{fig:vis_compare}. Compared to other methods, our CATANet can accurately restore clean edges with fewer artifacts because it captures similar textures from long ranges to supplement more long-range information, which indicates the superiority of our method. More visual examples can be found in the supplementary material.
\begin{table}
 \scriptsize
\caption{Ablation Study on IASA and IRCA. PSNR are calculated with a scale factor of 4.}
\vspace{-6mm}
\begin{center}
\resizebox{\linewidth}{!}{
\begin{tabular}{|cc|c|c|c|c|c|c|c|}
\hline
IASA&IRCA & Params & Multi-Adds&  Set5 & Set14 & B100 & Urban100& Manga109\\
\hline
\usym{2717}&\usym{2717}&366K &37.3G&32.26 & 28.63 &27.68 &26.46&30.81\\

\usym{2713}&\usym{2717}&511K &46.8G&32.47  &28.75  &27.75  &26.85 &31.24\\
\usym{2713}&\usym{2713}&  535K& 46.8G&
\textcolor{red}{32.58} & \textcolor{red}{28.90}  & \textcolor{red}{27.75} &  \textcolor{red}{26.87} & \textcolor{red}{31.31}\\
\hline
\end{tabular}}
\label{tab:iasa_irca}
\end{center}
\vspace{-7mm}
\end{table}
\begin{table}
\scriptsize
\caption{Ablation Study on different designs of TAB. PSNR are calculated with a scale factor of 2.}
\vspace{-6mm}
\begin{center}
\resizebox{\linewidth}{!}{
\begin{tabular}{|c|c|c|c|c|c|}
\hline
Method & Set5 & Set14 & B100 & Urban100& Manga109\\
\hline
Clustered Attention~\cite{vyas2020fast}&32.25&33.84&32.33&32.96&39.30\\
TCformer~\cite{zeng2022not}&38.06&33.87&32.32&32.90&39.17\\
NLSA~\cite{mei2021image}&37.67&33.29&31.96&31.22&37.78\\
\textbf{CATANet (ours)}&
\textcolor{red}{38.28} & \textcolor{red}{33.99}  & \textcolor{red}{32.37} &  \textcolor{red}{33.09} & \textcolor{red}{39.37}\\
\hline
\end{tabular}}
\label{tab:design}
\end{center}
\vspace{-7mm}
\end{table}
\begin{table}
\scriptsize
\caption{Ablation Study on fusion approach of IASA and IRCA. PSNR are calculated with a scale factor of 4.}
\vspace{-6mm}
\begin{center}
\resizebox{\linewidth}{!}{
\begin{tabular}{|l|c|c|c|c|c|}
\hline
Method & Params& Multi-Adds &  Set5 & Urban100& Manga109\\
\hline
concat& 548K&47.6G&32.49& 26.82&31.28\\
\textbf{add (ours)}& 535K& 46.8G&
\textcolor{red}{32.58} & \textcolor{red}{26.87} & \textcolor{red}{31.31}\\
\hline
\end{tabular}}
\label{table:fusion}
\end{center}
\vspace{-8mm}
\end{table}
\subsection{Ablation Study}
In this section, we conduct ablation studies to better understand and evaluate each component in the proposed CATANet. For a fair comparison, we implemented all experiments based on $\times4$ CATANet and trained them under the same settings. Besides, we set the input size as $3\times256\times256$ to compute Multi-Adds. \\
\textbf{Effects of IASA and  IRCA.} The \textbf{I}ntr\textbf{a}-Group \textbf{S}elf-\textbf{A}ttention (IASA) and \textbf{I}nte\textbf{r}-Group \textbf{C}ross-\textbf{A}ttention (IRCA) is a core component of CATANet to capture long-range dependencies for recovering damaged images. To evaluate the effectiveness of the proposed modules, we establish three models and compare their ability for image SR. \textbf{\textit{The first model (row 1)}} is the baseline model; we remove TAB and only adopt LRSA to process image features. To demonstrate the effectiveness of IASA, we propose \textbf{\textit{the second model (row 2)}} that includes an additional IASA branch on top of the first model. In \textbf{\textit{the third model (row 3)}}, we employ both IASA and IRCA simultaneously to show that IRCA can further enhance the utilization of long-range information. As shown in Tab.~\ref{tab:iasa_irca}, we can observe that with the help of IASA, \textbf{\textit{the second model (row 2)}} has a significant improvement on most datasets compared to \textbf{\textit{the first model (row 1)}}, with a maximum improvement of 0.17dB. With the joint help of IASA and IRCA, \textbf{\textit{the third model (row 3)}} once again has a significant improvement in all five datasets compared to \textbf{\textit{the second model (row 2)}}, with a maximum increase of 0.15dB. These are notable boosts in lightweight image SR.\\
\textbf{Effects of different designs of TAB.} To evaluate the effectiveness of our proposed Token Aggregation Block (TAB), we compare it with several other token aggregation methods in Tab.~\ref{tab:design}. Specifically, the Clustered Attention~\cite{vyas2020fast} replaces $\Q$ in standard attention with cluster centers and then performs attention and upsampling. However, upsampling is a coarse operation and cannot effectively restore important details. TCFormer~\cite{zeng2022not} is similar to Clustered Attention, with the main difference being that it uses KNN to merge neighboring pixels to obtain sparse $\Q$, while Clustered Attention uses k-means to achieve sparse $\Q$. NLSA~\cite{mei2021image} utilizes pixel hash values to group pixels with the same hash into one group and then performs standard attention within each group. On the one hand, hash-based methods may result in hash collisions, leading to inaccurate groupings. On the other hand, NLSA also enforces group size uniformity, which can lead to similar pixels being placed in adjacent blocks, a situation that NLSA does not address. We address this issue in IASA by allowing the $\Q$ of each group to attend to $\K$ and $\V$ of adjacent groups.\\
As shown, these methods actually obtains slightly poor performance than ours. The reason lies in component of TAB. Specifically, Content-Aware Token Aggregation module computes the similarity between tokens and token centers rather than relying on hashing, thereby enhancing the accuracy of grouping. IASA performs attention among content-similar tokens to achieve more refined information interaction, complementing the information from IRCA.\\
\textbf{Effects of Fusion Approach.}  We evaluate the impact of fusion methods for IASA and IRCA on performance. In the first model, we add the two output features together, while in the second model, we concatenate the two output features. As shown in Tab.~\ref{table:fusion}, the addition method achieves the best performance with a PSNR improvement of 0.03dB$\mathbf{\sim}$0.09dB while maintaining lower complexity. These experimental results indicate that adding the output features together is more efficient and lightweight than concatenating them.\\
\textbf{More analysis.} In the supplementary material, we provide more ablation studies and analysis.
\subsection{Visualization  Analysis} 
\begin{figure*}
    \centering
    
    \captionsetup{font={small}}
     \includegraphics[width=\linewidth]{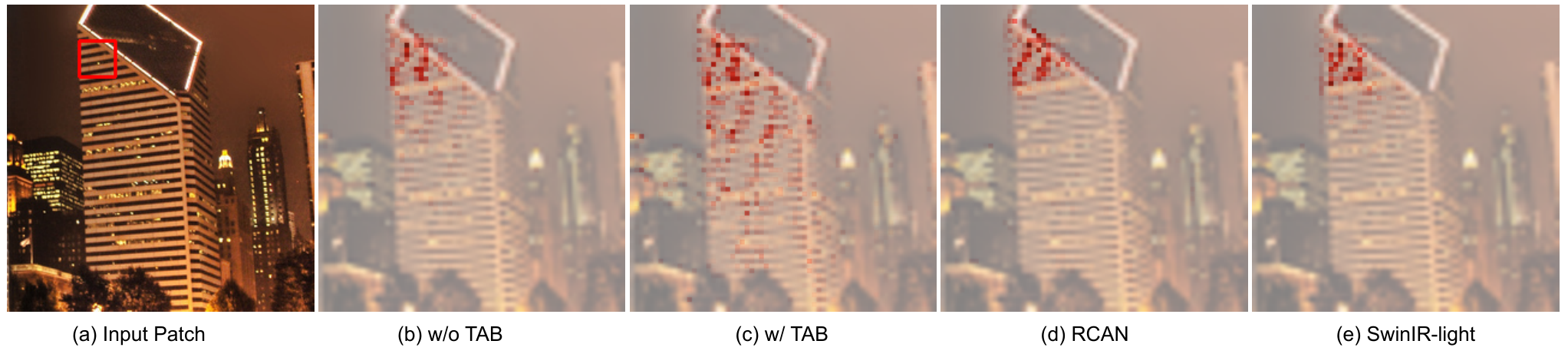}
    \vspace{-7mm}
\caption{LAM Comparison: the variant without TAB (w/o TAB), the full CATANet (w/ TAB), RCAN and SwinIR-light for $\times$4 SR.}
\label{fig:vis_lam}
\vspace{-2mm}
\end{figure*}

To better understand the improvement brought by the TAB, we utilize LAM \cite{gu2021interpreting} to visualize the effective receptive field of an input patch. As shown in Fig.~\ref{fig:vis_lam}, with the assistance of the TAB, our CATANet (w/ TAB) benefits from more useful long-range token information compared to the first model (w/o TAB). Additionally, we compared our model with the CNN-based method RCAN~\cite{Zhang_Li_Li_Wang_Zhong_Fu_2018} and the window-based attention method SwinIR-light~\cite{liang2021swinir}. The LAM map again demonstrates that the TAB captures more long-range information and achieves a larger receptive field. These experimental results further demonstrate the advantages of TAB.
\subsection{Model Size and Running Time Analyses}
\label{sec:mz_time}
\begin{table}
  \scriptsize
    \caption{Parameter, Multi-Adds, and Running Time comparison for scale factor $\times$4. CATANet-L(w/o $\mathcal{S}$) indicates CATANet-L without dividing token groups \(\mathcal{G}\) into subgroups \(\mathcal{S}\). The test input image size is 3$\times$256$\times$256. The best and second best results are colored with \textcolor{red}{red} and \textcolor{blue}{blue}, 
 respectively.}
 \vspace{-2mm}
  \centering
  \begin{tabular}{|c|c|c|c|}
\hline
Model & Params & Multi-Adds &  Time\\
\hline
SwinIR-light~\cite{liang2021swinir}&897K&60.3G&158.1ms\\
SRFormer-light~\cite{zhou2023srformer}&873K&56.5G&220.1ms\\
ATD-L~\cite{zhang2024transcending}&494K&30.0G&\textcolor{blue}{144.2ms}\\
SPIN~\cite{zhang2023lightweight}&555K&48.4G&435ms\\
CATANet-L(w/o $\mathcal{S}$)& 535K& 46.8G& 188ms\\
\textbf{CATANet-L (ours)}& 535K& 46.8G& \textcolor{red}{86ms}\\
\hline
  \end{tabular}

 \label{tab:time}
 \vspace{-3mm}
\end{table}
\begin{table}
\scriptsize
\caption{PSNR comparison between our CATANet-L/M/S and other lightweight methods for scale factor $\times 2$. $\dagger$ indicates being trained on DF2K~\cite{timofte2017ntire}.}
\vspace{-6mm}
\begin{center}
\resizebox{\linewidth}{!}{
\begin{tabular}{|c|c|c|c|c|c|}
\hline
Method &Params& Set5 & Set14 & B100 & Urban100\\
\hline
ATD-L~\cite{zhang2024transcending}&484K&38.23&33.94&32.33&32.95\\
SMFANet+~\cite{zheng2025smfanet}&480K&38.18&33.82&32.28&32.64\\
\textbf{CATANet-L (ours)}&477K& \textcolor{red}{38.28} & \textcolor{red}{33.99} & \textcolor{red}{32.37} & \textcolor{red}{33.09} \\
\hline
ATD-M~\cite{zhang2024transcending}&300K&38.14&33.78&32.26&32.58\\
DITN-Tiny~\cite{liu2023unfolding}&367K&38.00&33.70&32.16&32.08\\
\textbf{CATANet-M (ours)}&306K& \textcolor{red}{38.17} & \textcolor{red}{33.94} & \textcolor{red}{32.29} & \textcolor{red}{32.67} \\
\hline
ATD-S~\cite{zhang2024transcending}&229K&38.07&33.67&32.20&32.29\\
SAFMN~\cite{sun2023spatially}&228K&38.00&33.54&32.16&31.84\\
SeemoRe-T~\cite{zamfir2024details}$\dagger$&220K&38.06&33.65&32.23&32.22\\
\textbf{CATANet-S (ours)}&230K&\textcolor{red}{38.13} & \textcolor{red}{33.80} & \textcolor{red}{32.25} & \textcolor{red}{32.50} \\
\hline
\end{tabular}}
\label{tab:v_z}
\end{center}
\vspace{-0.8
cm}
\end{table}
To demonstrate the effectiveness and efficiency of CATANet, we design three variants with different model size (S/M/L) and evaluate their PSNR results and inference speed.\\
\textbf{Model Size} In Fig.~\ref{fig:demo} and Tab.~\ref{tab:v_z}, we compare the performance and complexity of our CATANet-L/M/S with other lightweight SR methods, where ATD-light~\cite{zhang2024transcending} is rescaled to similar model sizes as our three models denoted by ATD-S, ATD-M, and ATD-L. The results show that our CATANet-L/M/S achieves higher PSNR than other lightweight methods at each model size. Specifically, CATANet-M outperforms DITN-Tiny a large margin 0.59dB with fewer than 60K parameters and CATANet-S outperforms SeemoRe-T with a maximum PSNR improvement of 0.28 dB, even though SeemoRe-T was trained on a larger dataset.
\\
\textbf{Running Time} To reduce the accidental error, we run each model 100 times on one RTX 4090 GPU and calculate the average time as the final running time. As shown in the Tab.~\ref{tab:time}, our CATANet-L achieves the fastest peed. Compared to the window-based methods SwinIR-light~\cite{liang2021swinir} and SRFormer-light~\cite{zhou2023srformer}, our CATANet-L achieves approximately double their speed. Compared to SPIN, which relies on token aggregation, CATANet benefits from our proposed  efficient Content-Aware Token Aggregation, resulting in a significantly faster runtime, approximately five times faster than SPIN. Compared to ATD-L, which performs multiple types of attention in parallel, increasing the computational burden, our CATANet-L achieves approximately $1.5\times$ the speed. Additionally, we also compared CATANet-L and CATANet-L without dividing the token groups \(\mathcal{G}\) into subgroups \(\mathcal{S}\), and CATANet-L is nearly 2 times faster.
\section{Conclusion}
In this paper, we propose a novel Lightweight Image Super-Resolution network called CATANet, which leverages token centers to aggregation content-similar tokens into content-aware regions. Specifically, the core component of CATANet is Token-Aggregation Block. Token-Aggregation Block is mainly comprised of Content-Aware Token Aggregation (CATA), Intra-Group Self-Attention, and Inter-Group Cross-Attention. CATA module efficiently aggregate similar tokens together, forming content-Aware regions. Intra-Group Self-Attention is responsible for fine-grained long-range information interaction within content-aware regions. At the same time, we introduce Inter-Group Cross-Attention, which applies cross-attention between each group and token centers to further enhance global information interaction. We have presented extensive experimental results on various benchmark datasets, and our method has achieved superior performance while maintaining high inference efficiency in Lightweight Image Super Resolution.

\section*{Acknowledgments}
This work was supported by the National Natural Science Foundation of China (Grant No. 62402211) and the Natural Science Foundation of Jiangsu Province (Grant No. BK20241248). The authors would also like to thank the support from the Collaborative Innovation Center of Novel Software Technology and Industrialization, Jiangsu, China.
{
    \small
    \bibliographystyle{ieeenat_fullname}
    \bibliography{main}
}
\end{document}